\documentclass[conference]{IEEEtran}
\IEEEoverridecommandlockouts

\usepackage[numbers]{natbib}
\usepackage{amsmath,amssymb,amsfonts}
\usepackage{algorithmic}
\usepackage{graphicx}
\usepackage{textcomp}
\usepackage{xcolor}
\usepackage{enumitem}
\usepackage{listings}
\usepackage{url}

\def\BibTeX{{\rm B\kern-.05em{\sc i\kern-.025em b}\kern-.08em
    T\kern-.1667em\lower.7ex\hbox{E}\kern-.125emX}}
\begin{document}

\lstset{
    basicstyle=\ttfamily\small,
}

\title{An Artificial Intelligence Driven Semantic Similarity-Based Pipeline for Rapid Literature Review\\
}

\author{\IEEEauthorblockN{Abhiyan Dhakal}
\IEEEauthorblockA{\textit{Kathmandu University} \\
Dhulikhel, Nepal \\
itsabhiyandhakal@gmail.com}
\and
\IEEEauthorblockN{Kausik Paudel}
\IEEEauthorblockA{\textit{Kathmandu University} \\
Dhulikhel, Nepal \\
kausikpaudel@gmail.com}
\and
\IEEEauthorblockN{\textsuperscript{*}Sanjog Sigdel}
\IEEEauthorblockA{\textit{Kathmandu University} \\
Dhulikhel, Nepal \\
sanjog.sigdel@ku.edu.np}
\thanks{\textsuperscript{*}Corresponding author: Sanjog Sigdel (sanjog.sigdel@ku.edu.np).}
}

\maketitle

\sloppy

\begin{abstract}

We propose an automated pipeline for performing literature reviews using semantic similarity. Unlike traditional systematic review systems or optimization-based methods, this work emphasizes minimal overhead and high relevance by using transformer-based embeddings and cosine similarity. By providing a paper's title and abstract, it generates relevant keywords, fetches relevant papers from open-access repositories (e.g., ArXiv), and ranks them based on their semantic closeness to the input. Three embedding models: TF-IDF, all-MiniLM-L6-v2, and Specter2 were evaluated. While TF-IDF struggled with capturing deeper semantic meaning, all-MiniLM-L6-v2 provided broader conceptual coverage. Specter2, specifically fine-tuned for scientific texts, exhibited score saturation in similarity scores. A statistical thresholding approach is then applied to filter relevant papers, enabling an effective literature review pipeline. Despite the absence of heuristic feedback or ground-truth relevance labels, the proposed system shows promise as a scalable and practical tool for preliminary research and exploratory analysis.

\end{abstract}

\section{Introduction}
Conducting a literature review is an important step in any research project. By examining existing studies, researchers understand the progress that has been made in their field. Literature reviews summarizes the current state of research, and identify gaps providing insights to the direction of future studies. 

Traditional systematic literature review (SLR) methods are mostly time consuming and require extensive manual efforts to select papers, filter them, and organize relevant papers. Recent automated approaches, such as Swarm SLR \cite{b1} has attempted to address these challenges by introducing structured workflow and visualization for systematic reviews, however, it requires extensive configuration, user's expertise for proper parameter tuning and possesses challenges to perform quick, exploratory literature reviews on the provided context. 

Considering the research gap, we propose AutoLit, an automated pipeline designed to perform literature reviews using semantic similarity. AutoLit operates on minimal input. It requires the title and abstract of the research paper, and generates a list of relevant papers based on transformer-based embeddings and cosine similarity. The pipeline automatically generates keywords from the input, retrieves papers from open-access repositories like arXiv \cite{b2}, and filters them based on semantic similarity using a customizable threshold.

This approach is useful for early-stage research and exploratory analysis, focusing on quick and relevant results rather than complete coverage. It minimizes manual intervention, and thus provides an efficient alternative to existing SLR systems for efficient literature discovery and reviews.

\section{Related Works}


With the development and innovation of Large Language Models (LLMs), semantic embeddings and workflow optimization techniques, automating the process of literature review has seen significant progress in the recent years. In replacement of traditional SLR,  Wittenborg et al. \cite{b1} introduced SWARM-SLR that has effectively structured the systematic literature review process by defining clear requirements, establishing a modular workflow and creating a framework to evaluate the supporting tools. This has had impact in the early stages of research, such as searching other relevant papers and initial data processing. However, SWARM-SLR lacks the actual process of writing literature review.  

In contrast to SWARM-SLR, Tang et al.~\cite{b3} has explored the feasibility of LLMs in automating various processes of literature reviews such as reference generation, abstraction composition and complete review writing. Their evaluation showed that while LLMs showed impressive results, they often suffer from hallucination in references, favoring the tasks most cited. Additionally, generating incomplete author lists, and variable performance across various disciplines were major limitations. This is partly addressed by Li et al.~\cite{b4} through their system ChatCite, which combines human workflow guidance within the LLM agent. It is a semi-automated system that generates comparative summaries between studies, which was not addressed by prior LLM based approaches. However, it relies on curated datasets and thus may not generalize summaries without further adaptation.

Reimers and Gurevych~\cite{b5} introduced Sentence Bert (SBERT) for semantic similarity, which has become a baseline in many automated pipelines largely due to its efficient and accurate generation of sentence embeddings. SBERT enables precise comparison using cosine similarity. This significantly improves tasks such as relevant literature retrieval, ranking, and clustering which is useful for understanding the closeness of different papers. 

\begin{figure*}[t]
    \setlength{\fboxsep}{12pt}   
    \setlength{\fboxrule}{1pt}  
    \centering
    \fbox{\includegraphics[width=.8\linewidth]{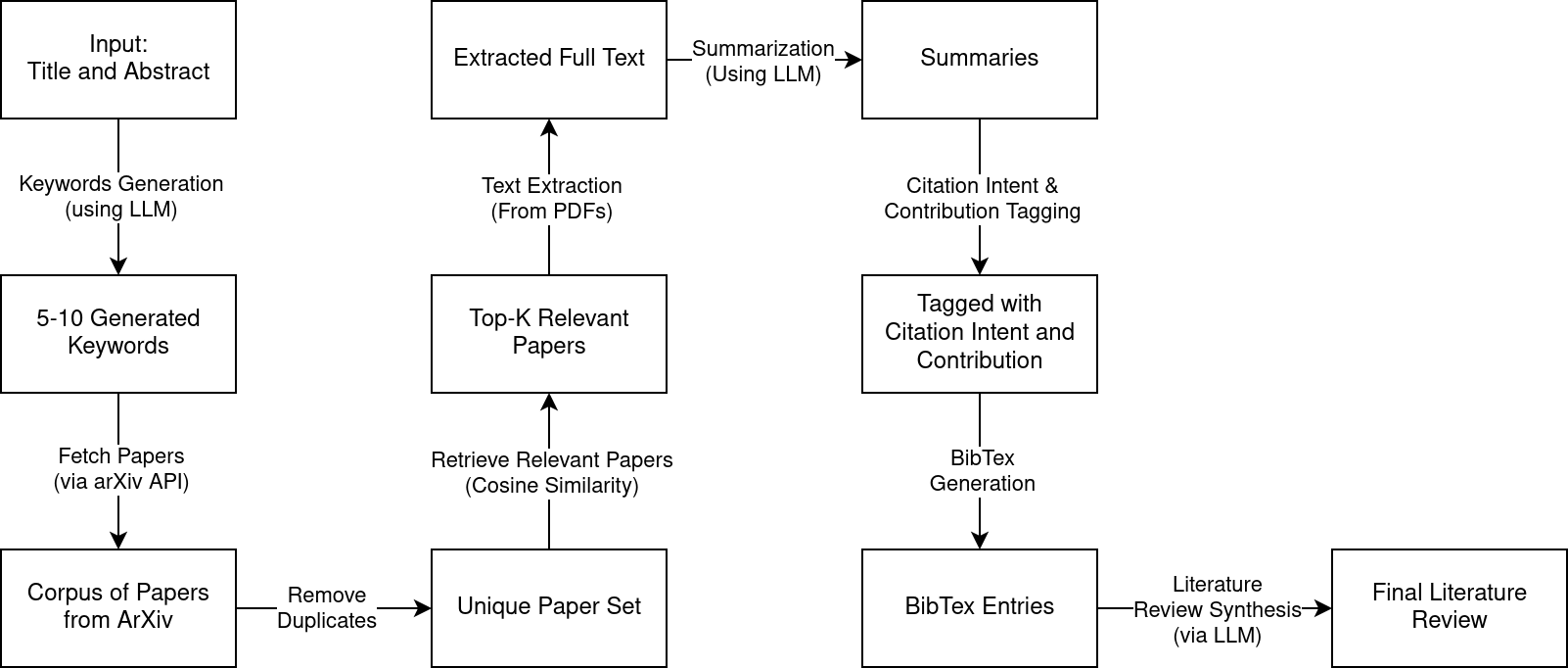}}
    \caption{Pipeline Overview}
    \label{fig:pipeline-overview}
\end{figure*}

Kasanishi et al.~\cite{b6} proposed SciReviewGen, a large-scale dataset and framework designed for automatic generation of literature reviews by summarizing multiple scientific papers using transformer-based models like Fusion-in-Decoder~\cite{b7} extended for literature review generation. It focuses on producing summaries but still faces challenges like hallucinations and missing detailed information.

Additionally, Wang et al.~\cite{b8} evaluate the capability of ChatGPT in generating Boolean queries for systematic literature searches. Their findings showed that query refinement using ChatGPT can improve the recall and F1 scores. The refined query after single prompting resulted an F1 score of 0.0772 while guided prompting increased the F1 score to 0.5171 which is a significant rise. Also the highest recall they were able to achieve was 0.9128. This indicates that combining base queries with ChatGPT refinement can generate effective Boolen queries suitable for rapid literature reviews where time constraints matter. Although LLMs can approximate human like query construction, they fall short in encoding complex Boolean logic effectively.

Despite these advances, many existing systems are based on multi-stage workflows, which can significantly hamper the performance of the system. In contrast, semantic pipelines that prioritize minimal input and provide relevant results would be beneficial for streamlining the literature review process. This approach addresses the current gap by focusing on semantic similarity-based classification and filtering of open-access repositories such as arXiv. It emphasizes transformer-based embeddings and cosine similarity for semantic search that enables the extraction of a highly relevant set of papers while performing relevant literature review of the given paper.

\section{Methodology}

This section outlines the pipeline designed to automate the literature review process. An overview of the pipeline is illustrated in Figure~\ref{fig:pipeline-overview}. The pipeline consisted of seven key stages, each designed to efficiently and systematically process scholarly articles relevant to the research topic.

\subsection{Keyword Generation}

\begin{figure}[h]
    \setlength{\fboxsep}{12pt}   
    \setlength{\fboxrule}{1pt}  
    \centering
    \fbox{\includegraphics[width=0.5\linewidth]{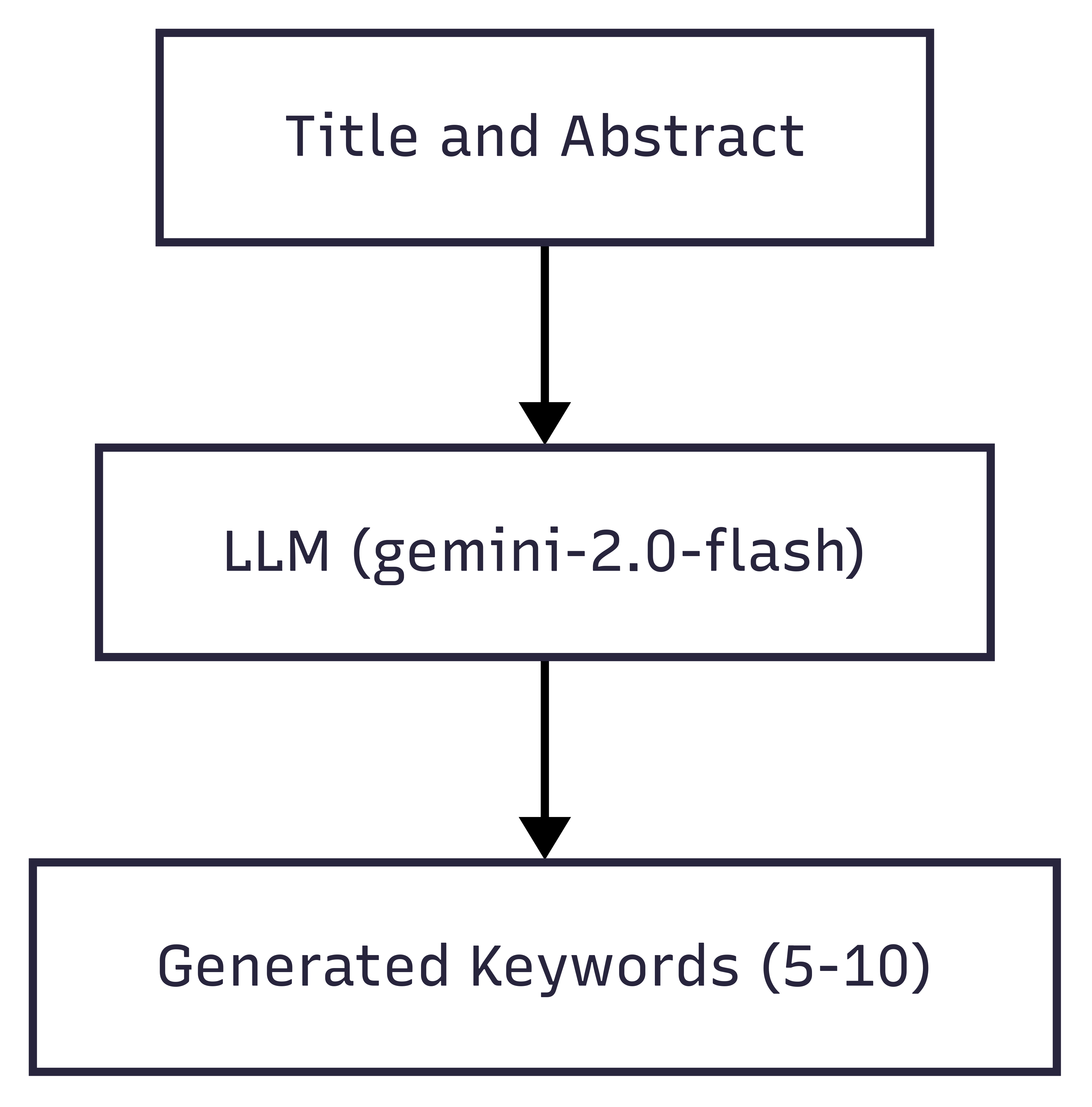}}
    \caption{Keyword Generation Processs}
    \label{fig:keyword-gen}
\end{figure}

The initial stage involved generating a set of relevant keywords to guide the literature search. Given the title and abstract of the target paper, 5 to 10 keywords were generated using a LLM, \texttt{gemini-2.0-flash}~\cite{b9}, which recent studies have shown to be effective for zero-shot keyphrase extraction~\cite{b10}, to capture both explicit and semantically related terms, as illustrated in Figure~\ref{fig:keyword-gen}.

\subsection{Fetch Papers}

For each keyword generated, a corpus of a maximum of 20 relevant papers was fetched from arXiv via keyword-based search using the arXiv API. Since the papers were being fetched using different keywords, there were some duplicates. Those duplicate papers were filtered out on the basis of their metadata ensuring that only unique papers remained in the corpus.

\subsection{Retrieve Relevant Papers}

To refine the selection of the papers obtained, a semantic similarity assessment was performed. The sentence transformer model \texttt{all-MiniLM-L6-v2}~\cite{b11} was used to generate 384-dimensional dense vector embeddings for the input query (derived from keywords) and the title and abstract of the papers obtained. The cosine similarities between the embeddings of the fetched papers and that of the input query were then calculated.

A statistical thresholding method based on the interquartile range (IQR) of the similarity scores distribution was applied to filter the papers.
The IQR method is a robust, distribution-free approach for outlier detection that does not assume normality of the data \cite{b12,b13}. In line with recent NLP work \cite{b14}, we adopt a conservative upper bound of $Q3 + 0.5 \times \mathrm{IQR}$ to select only the highest-scoring papers, ensuring high precision while retaining adaptability across different similarity score distributions.

\begin{figure}[h]
    \centering
    \setlength{\fboxsep}{12pt}   
    \setlength{\fboxrule}{1pt}  
    \fbox{\includegraphics[width=0.89\linewidth]{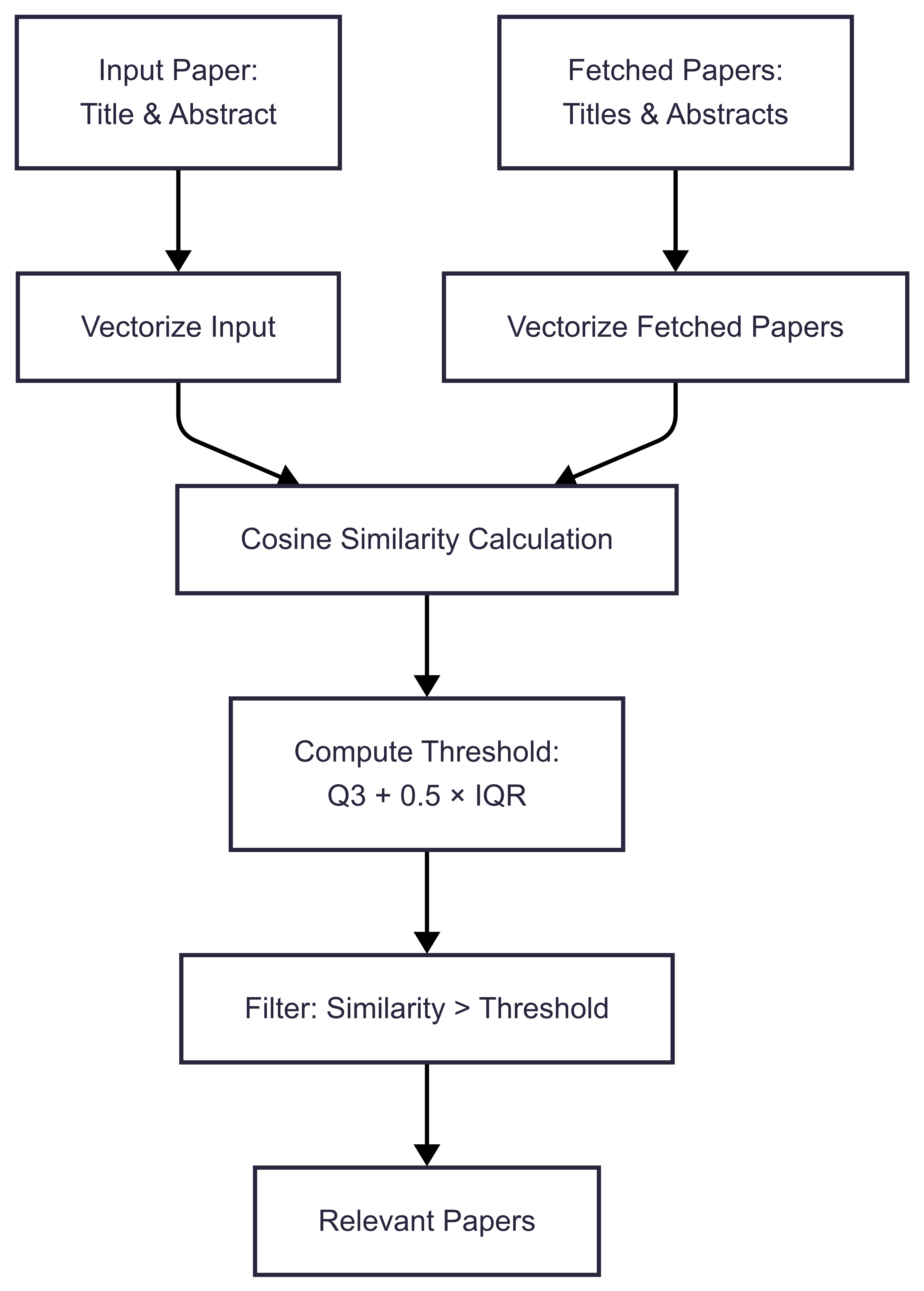}}
    \caption{Relevant Papers Retrieval Processs}
    \label{fig:similarity-filter}
\end{figure}

This adaptive thresholding approach accounted for the natural variance and skew in the similarity scores, resulting in a more reliable filtering of semantically relevant papers, as illustrated in Figure~\ref{fig:similarity-filter}.

For comparison, we also evaluated other embedding models, including Specter2~\cite{b15}, which is specifically trained on scientific citation data, and a traditional TF-IDF with cosine similarity baseline. Specter2 produced tightly clustered high similarity scores that required stricter thresholding to avoid false positives, while TF-IDF offered greater precision but lacked semantic coverage, often missing conceptually related papers. Based on this evaluation, \texttt{all-MiniLM-L6-v2} was selected as a balanced choice for semantic retrieval in this pipeline.

\subsection{Text Extraction from PDF}

For each filtered paper, either the abstract or the full PDF was processed to extract structured textual content. We used \texttt{PyMuPDF}~\cite{b16} library for full text extraction when PDFs were available. To structure the content of each paper, regular expression (regex) patterns were defined to identify and delineate different sections: Introduction, Methodology, Results, and Conclusion. The defined patterns were as follows:
\begin{itemize}
    \item \textbf{Abstract}: The regex pattern \texttt{(?i)\textbackslash babstract\textbackslash b} was used to locate the abstract section of the paper.
    \item \textbf{Introduction}: The pattern \texttt{(?i)\textbackslash bintroduction\textbackslash b} was applied to identify the introduction.
    \item \textbf{Methods}: To capture the methodology section, the pattern \texttt{(?i)\textbackslash b(methodology|methods|approach)\textbackslash b} was used, accounting for various terminologies that might describe this section.
    \item \textbf{Results}: To capture the results section, the pattern \texttt{(?i)\textbackslash b(results|findings|experiments)\textbackslash b} was employed, which might also be referred to as findings or experiments.
    \item \textbf{Conclusion}: The regex pattern \texttt{(?i)\textbackslash b(conclusion|discussion|summary)\textbackslash b} was defined to capture the conclusion, which might also appear under the names "discussion" or "summary" in some papers.
\end{itemize}

\subsection{Summarization}

The extracted text from each section of the relevant papers was then summarized using the \texttt{gemini-2.0-flash} LLM. To ensure consistent information extraction, the LLM was instructed to structure the summaries under the \texttt{summary} key, which contained four distinct categories: \texttt{problem\_statement}, \texttt{methodology}, \texttt{key\_findings}, and \texttt{conclusion\_recommendations} (see Listing~\ref{lst:summary-json}). Each of these categories was designed to hold a list of bullet points summarizing the respective information extracted from the paper.

\begin{lstlisting}[caption={Example of structured summary JSON}, label={lst:summary-json}]
{
  "summary": {
    "problem_statement": [
      "Bullet point 1",
      "Bullet point 2"
    ],
    "methodology": [
      "Bullet point 1",
      "Bullet point 2"
    ],
    "key_findings": [
      "Bullet point 1",
      "Bullet point 2"
    ],
    "conclusion_recommendations": [
      "Bullet point 1",
      "Bullet point 2"
    ]
  }
}
\end{lstlisting}

Specifically, under the \texttt{problem\_statement} category, the LLM would list the identified research problems or gaps addressed in the paper as individual bullet points. The \texttt{methodology} category would contain bullet points describing the research approach, methods, or techniques employed by the authors. The \texttt{key\_findings} category would present the most significant results or discoveries reported in the paper in a bulleted format. Finally, the \texttt{conclusion\_recommendations} category would list the main conclusions drawn by the authors, along with any recommendations for future work or applications of their findings.

\subsection{Citation Intent and Contribution Tagging}

To further analyze the nature and contributions of each selected paper, the Gemini LLM was employed to perform two key tasks:
\begin{enumerate}[label=(\roman*)]
  \item \textbf{Citation intent tagging}, following a predefined taxonomy: \textit{Background}, \textit{Comparison}, \textit{Extension}, \textit{Criticism}, \textit{Application}, \textit{Future Work}, and \textit{Other}.
  \item \textbf{Contribution classification}, assigning each paper to one of: \textit{Dataset}, \textit{Algorithm}, \textit{Framework}, \textit{Review}, \textit{Benchmark}, \textit{Survey}, \textit{System}, \textit{Theoretical Analysis}, or \textit{Other}.
\end{enumerate}

This approach follows prior research on citation function classification and contribution categorization in scientific literature~\cite{b17, b18, b19} and recent advances employing LLMs for automated citation intent and contribution tagging.

\subsection{BibTex Entry and Literature Review Generation}

\begin{figure}[h]
    \centering
    \setlength{\fboxsep}{12pt}   
    \setlength{\fboxrule}{1pt}  
    \fbox{\includegraphics[width=.89\linewidth]{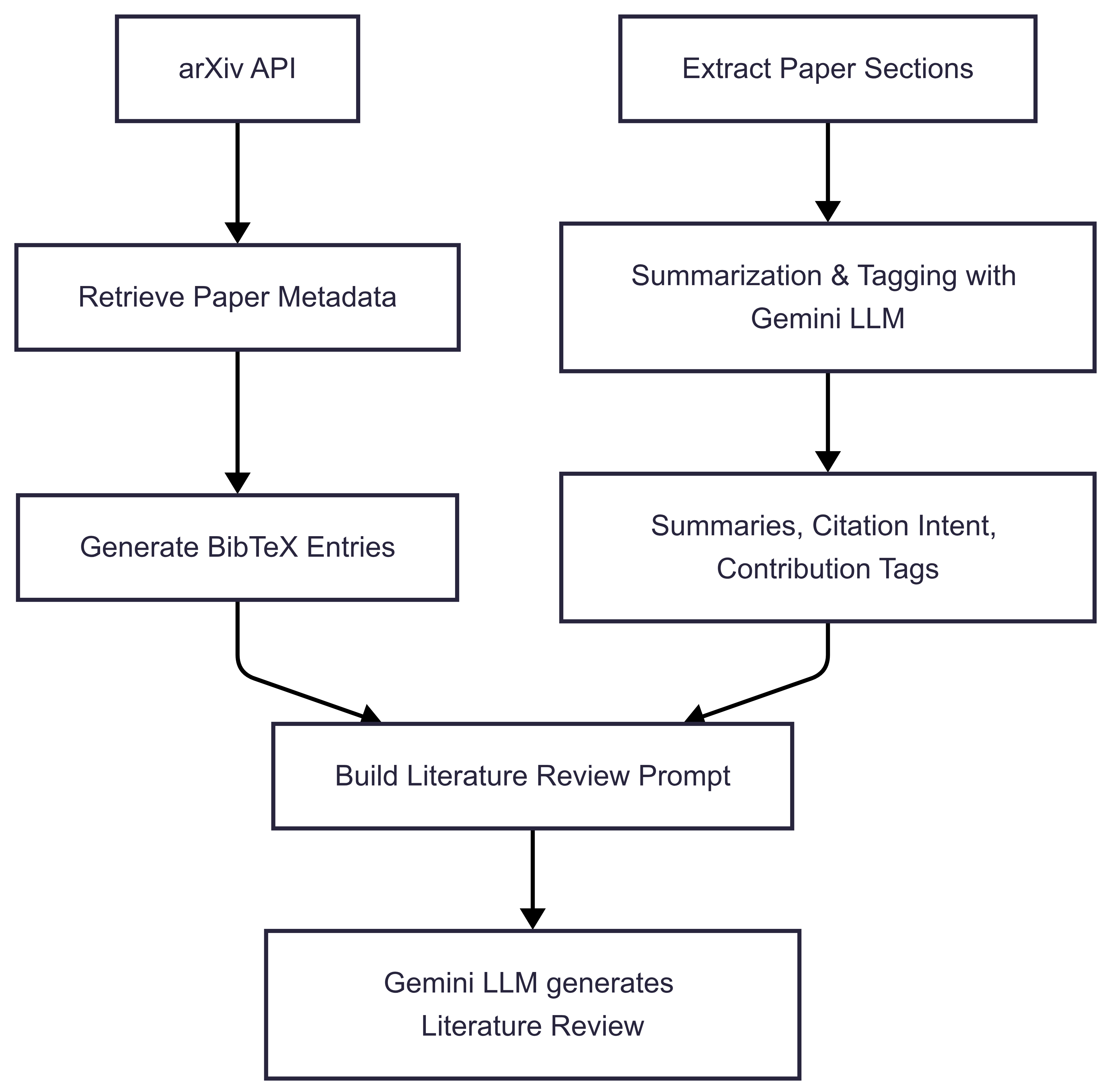}}
    \caption{Litereature Review Synthesis}
    \label{fig:lit-synthesis}
\end{figure}

The overall pipeline for generating the literature review is shown in Figure~\ref{fig:lit-synthesis}. It begins with metadata retrieval from the arXiv API, followed by BibTeX entry generation, paper summarization and tagging using the Gemini LLM, and finally synthesis of the literature review paragraph based on these structured inputs.

The metadata of the relevant papers, retrieved from the arXiv API in earlier stages, was utilized to generate BibTeX entries for each paper. This ensured proper citation formatting for any inclusion of these sources in the final paper.

For the generation of the literature review itself, the summaries, citation intent tags, and contribution type tags, previously obtained using the \texttt{gemini-2.0-flash} LLM, served as the primary input. The Gemini model was instructed to synthesize this information into a comprehensive overview of the existing literature related to the research topic. This step aimed to provide a high-level understanding of the current state-of-the-art and key contributions in the field, leveraging the automated analysis performed in the preceding stages of the pipeline.

\section{Evaluation}

To evaluate the effectiveness of embedding models in filtering relevant scientific papers for automated literature review, we experimented with three distinct approaches for research paper similarity:

\begin{itemize}
    \item \textbf{TF-IDF}: Classical lexical vectorization.
    \item \textbf{all-MiniLM-L6-v2}: General-purpose transformer embeddings.
    \item \textbf{Specter2}: Scientifically tuned transformer embeddings.
\end{itemize}

A total of 178 papers were fetched from arXiv API. Each input paper's vector was compared to fetched candidates using cosine similarity. Table~\ref{tab:comparative_summary} summarizes the distributions.

\begin{table*}[h]
    \centering
    \caption{Comparative Summary of Embedding Models and Their Performance}
    \begin{tabular}{lccccc}
        \hline
        \textbf{Embedding Model} & \textbf{Threshold} & \textbf{Skewness} & \textbf{Value Range} & \textbf{\# Retrieved Papers} \\
        \hline
        TF-IDF & 0.204 & 0.622 & [0.010, 0.294] & 19 \\
        all-MiniLM-L6-v2 & 0.659 & 0.390 & [0.070, 0.804] & 20 \\
        Specter2 & 0.924 & $-$0.963 & [0.756, 0.945] & 11 \\
        \hline
    \end{tabular}
    \label{tab:comparative_summary}
\end{table*}

\begin{figure*}[h]
    \centering
    \includegraphics[width=0.95\linewidth]{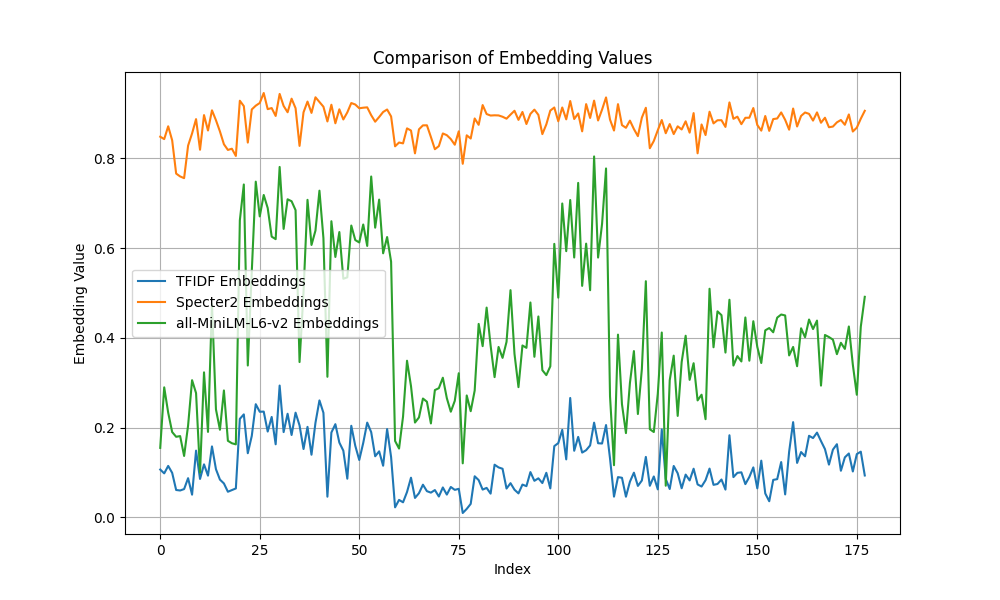}
    \caption{Line plot showing cosine similarity scores across models}
    \label{fig:similarity_lineplot}
\end{figure*}

\begin{figure*}[h]
    \centering
    \includegraphics[width=0.95\linewidth]{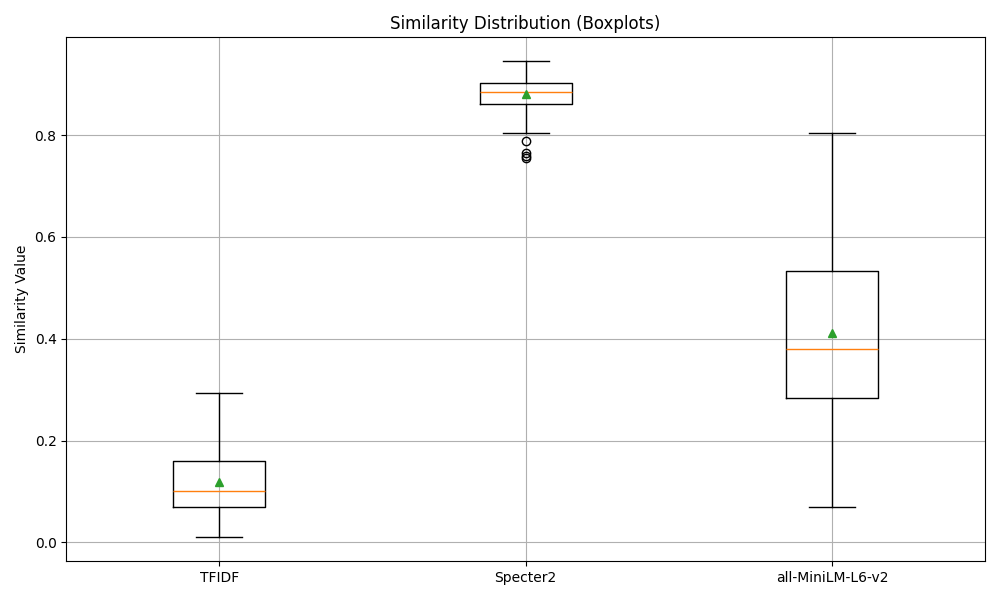}
    \caption{Similarity distribution visualization of models using box plots}
    \label{fig:similarity_boxplot}
\end{figure*}

\subsection{TF-IDF}

TF-IDF is a purely syntactic model, it lacks semantic encoding capabilities. With a calculated threshold of 0.204, it retained 19 papers. Although 19 articles were filtered, the cosine similarity distribution was highly positive and skewed toward the range [0.010, 0.294], indicating generally low similarity scores across the corpus, which is illustrated in Figure~\ref{fig:similarity_lineplot} and Figure~\ref{fig:similarity_boxplot}. Also, the model failed to capture the semantic meaning of the input paper even among the selected relevant papers above the threshold. Thus, TF-IDF excels in high lexical overlap but misses conceptual matches.

\subsection{all-MiniLM-L6-v2}

The all-MiniLM-L6-v2 model is a general-purpose semantic embedding model. With a calculated threshold set at 0.659, it retrieved 20 papers. The cosine similarity scores was slightly positively skewed in the range [0.070, 0.804] with similarity score of 0.390. Performance reflects a moderately spread distribution of similarity scores. Figure~\ref{fig:similarity_lineplot} and Figure~\ref{fig:similarity_boxplot} shows that this model provided convincing separations between relevant and less relevant documents compared to both TF-IDF and specter2. Unlike TF-IDF, this model captured semantic relations, enabling retrieval of relevant papers even when exact terminology differed.

\subsection{Specter2}

Specter2 is a fine-tuned model explicitly for scientific document embeddings. As a result, it achieved the highest absolute similarity scores. A threshold of 0.924 resulted in the retrieval of 11 papers, with the score distribution skewed in the range [0.756, 0.945] with a negative skewness value of 0.963. Specter2 effectively captures both semantics and domain-specific knowledge. However, with our threshold set at 0.924, the distribution exhibited higher density near 1.0. All the values were clustered at a range of 0.75 to 0.95 requiring an extremely precise threshold, which might not always be the case for varied input papers. This can result in false positive values.

\section{Limitations}

\subsection{Lack of Ground-Truth Relevance Labels}
The current system operates without any human-annotated or expert-verified relevance labels. This makes it challenging to assess whether the papers retrieved by the semantic similarity filter are truly relevant to the input query. In the absence of proper benchmarks, only indirect metrics such as similarity score distributions can be used to approximate relevance. This may not always reflect actual correctness.

\subsection{Generalization Constraints of Embedding Models}

While models like Specter2 are trained on scientific text, general-purpose transformers such as \texttt{all-MiniLM-L6-v2} are optimized for broad coverage across diverse domains rather than the scholarly literature specifically. Prior studies have shown that such models often underperform in specialized domains, as they may not capture domain-specific terminology, citation context, or methodological nuances with the same fidelity as models trained on in-domain corpora \cite{b20, b21, b22}. This mismatch can lead to embeddings that overlook subtle but important conceptual relationships, reducing retrieval precision in fields with specialized jargon or structured discourse, such as medicine, chemistry, or legal research.

\subsection{Heuristic Thresholding Method}
The thresholding method employed, using the third quartile plus half the interquartile range (\( Q3 + 0.5 \cdot IQR \)) is a statistical heuristic designed to filter out low-similarity papers. However, this approach does not guarantee semantic relevance. It may exclude important but lexically distant papers and include irrelevant ones that happen to lie within the upper tail of the similarity distribution. The threshold's effectiveness is also dependent on the characteristics of the embedding model and the corpus.

\subsection{Lack of Qualitative Evaluation}
Evaluation of the retrieved papers depends entirely on quantitative metrics such as cosine similarity distributions, mean scores, and standard deviations. There is no qualitative assessment of the relevance of the chosen papers in terms of their content or research methods. Conducting such studies would require curated gold-standard datasets or involvement of subject matter experts, which was not feasible within the current resource. Without involvement of human reviewers or case studies, it remains uncertain how well the system supports real-world tasks like literature reviews or knowledge discovery.

\section{Conclusion}

This work presents an automated pipeline for conducting targeted literature reviews using semantic similarity. By using transformer-based embeddings and large language models, it efficiently retrieves, filters, and summarizes academic literature with minimal human intervention. By integrating TF-IDF, all-MiniLM-L6-v2, and Specter2 embeddings, the system evaluated and filtered 178 papers based on their relevance to a given query derived from title and abstract prompts.

TF-IDF filters documents based on exact text matches, resulting in loss of semantic meaning. all-MiniLM-L6-v2 provides a balanced performance but lacks domain-specific tuning. Specter2 is best aligned with scientific language, however suffered from similarity saturation, due to rigid threshold calibration.

In future work, we plan to focus on implementing adaptive thresholding to dynamically adjust based on score distribution characteristics. Incorporating human-in-the-loop evaluation will provide qualitative insights and a ground-truth basis for refining automated decisions. Re-ranking techniques will be explored to help improve the precision and contextual relevance of the retrieved papers.

Beyond thresholding and ranking improvements, domain adaptation and model fine-tuning (e.g., on specialized corpora) will be investigated to better capture field-specific semantics. Integration with citation network data sources such as Semantic Scholar \cite{b23} can enable network-based relevance scoring (e.g., PageRank-style importance measures, co-citation patterns), which may enhance both retrieval accuracy and literature synthesis. Finally, optimizing query formulation techniques will be considered to increase recall while maintaining high relevance.

This work has shown the potential to reduce manual overhead in writing literature reviews, while maintaining relevance on topics, proving it is an scalable and accessible alternative to traditional SLR systems.

\end{document}